\documentclass[letterpaper, conference]{ieeeconf}  

\IEEEoverridecommandlockouts                              

\usepackage{graphicx}
\usepackage{subcaption}
\usepackage{gensymb}
\usepackage{hyperref}

\title{\LARGE \bf
Employing Laban Shape for Generating Emotionally and Functionally Expressive Trajectories in Robotic Manipulators
}

\author{Srikrishna Bangalore Raghu$^{\ast,1}$, Clare Lohrmann$^1$, Akshay Bakshi$^1$,\\ Jennifer Kim$^1$, Jose Caraveo Herrera$^1$, Bradley Hayes$^1$, and Alessandro Roncone$^1$
\thanks{This work was supported by the Army Research Laboratory under grant number W911NF-21-2-02905.}
\thanks{$^{\ast}$ Corresponding author. Email: srba2850@colorado.edu.}
\thanks{$^1$ Authors are with the Department of Computer Science, University of Colorado Boulder,
Boulder, CO, USA. Email: {first.last}@colorado.edu.}%
}

\begin{document}

\maketitle
\thispagestyle{empty}
\pagestyle{empty}

\begin{abstract}
Successful human-robot collaboration depends on cohesive communication and a precise understanding of the robot’s abilities, goals, and constraints. While robotic manipulators offer high precision, versatility, and productivity, they exhibit expressionless and monotonous motions that conceal the robot's intention, resulting in a lack of efficiency and transparency with humans. In this work, we use Laban notation, a dance annotation language, to enable robotic manipulators to generate trajectories with \textsl{functional expressivity}, where the robot uses nonverbal cues to communicate its abilities and the likelihood of succeeding at its task. We achieve this by introducing two novel variants of Hesitant expressive motion (Spoke-Like and Arc-Like). We also enhance the emotional expressivity of four existing emotive trajectories (Happy, Sad, Shy, and Angry) by augmenting Laban Effort usage with Laban Shape. The functionally expressive motions are validated via a human-subjects study, where participants equate both variants of Hesitant motion with reduced robot competency. The enhanced emotive trajectories are shown to be viewed as distinct emotions using the Valence-Arousal-Dominance (VAD) spectrum, corroborating the usage of Laban Shape.

\end{abstract}

\section{INTRODUCTION}
With the growing popularity of robots and their increased deployment in the real world, it has become increasingly important to ensure their successful collaboration
with humans. Effective human-robot collaboration relies on clear communication and an accurate understanding of the robot's mental model, which refers to the humans' understanding of the robots' capabilities, intentions, and limitations \cite{tabrez2020survey}. Even though verbal communication is the most common means of conveying information, humans rely on nonverbal cues during regular interactions \cite{burgoon2011nonverbal}, and deploying them on robots can bolster team cohesion and alignment \cite{nonverbal}.\\ 
\begin{figure}[hbt!]
\centering
\includegraphics[width=\linewidth]{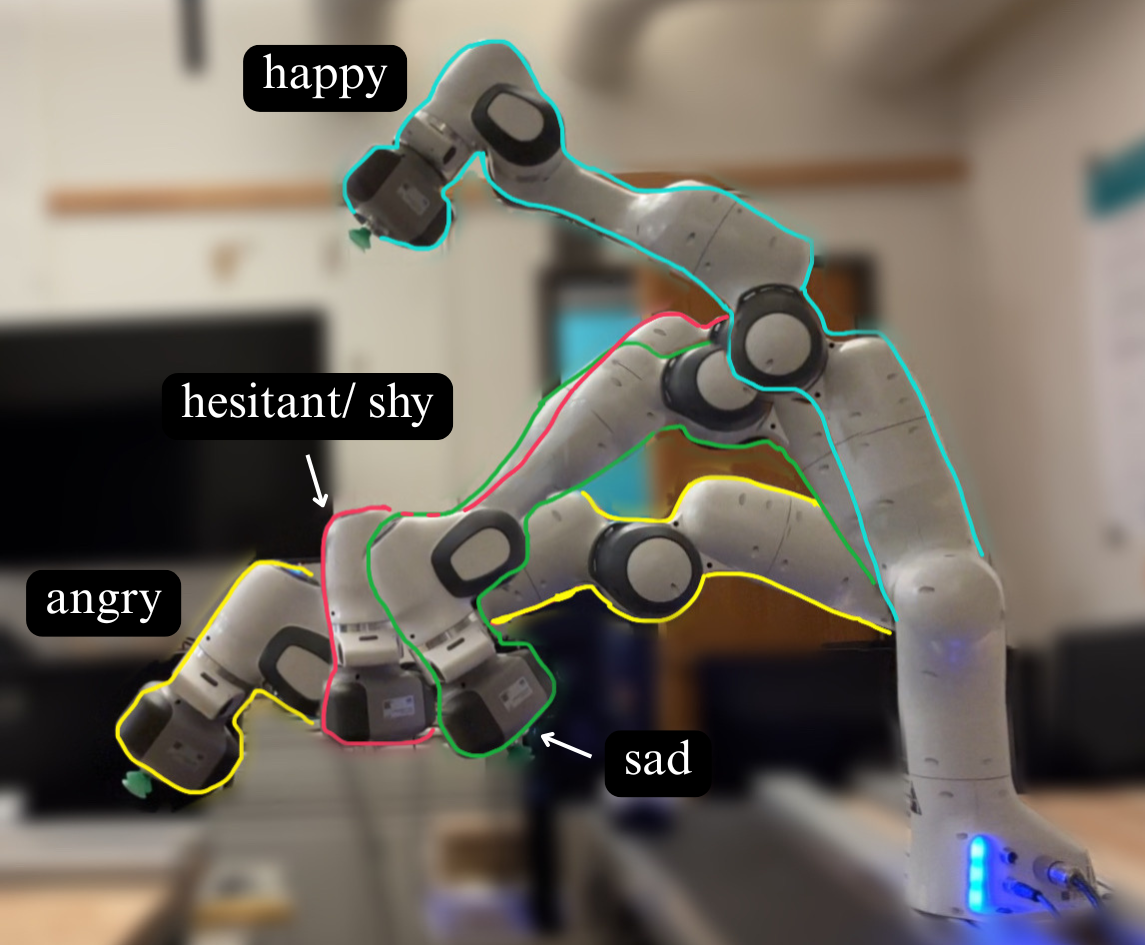}
\caption{A representation of all 4 Laban Shape Forms, which refers to the static shape the robot assumes throughout the expressive trajectory. The shape used for both variants of Hesitant/Uncertain motion (Spoke-Like and Arc-Like) is the same as the shape used in the Shy trajectory.}
\label{fig:all_shapes_together}
\end{figure}
Within the domain of expressive robotics, a vital component is \textsl{functional expressivity} \cite{mataric}. Functional accounts of emotion theory suggest that each emotion type serves as a social signal to enhance communication \cite{functional_new}. In this work, \textsl{we define \textbf{functional expressivity} as the robot's ability to communicate its functional capabilities and limitations}. In prior work, researchers have explored this concept through the use of visual cues such as arrows and navigation points in an augmented reality setting \cite{functionalexpressivity1}, and a combination of blinkers and beacons in a real-world setting \cite{functionalexpressivity2}. However, in addition to the above methods, \textsl{functional expressivity} could possibly be realized by specific nonverbal cues in the robot's motion. In pursuit of exploring this direction of research, we frame "Uncertainty" as the functional capability of the robot to express the unreliability and riskiness of successfully completing its task, and "Hesitancy" as the functional expression, i.e. what is used to express uncertainty.\\ 
For generating expressive motion in robots, an established approach is the Laban Movement Analysis (LMA) framework. LMA was introduced by Rudolf Laban in the 1940s for analyzing human movement and expression \cite{laban1975modern} and was later formalized by Irmgard Bartenieff \cite{bartenieff2013body}. LMA classifies movement into four categories that represent various aspects of motion: Body, Effort, Shape, and Space. It has been extensively applied to a diverse set of domains such as dance, technology, and psychology \cite{bernstein2015laban, psych, tsachor2017somatic}. Within computer science, it has been employed to design human detection algorithms \cite{bernardet2019assessing} and to enable expressiveness in robots \cite{differential_drive_intro, uav_intro, humanoid1, manipulators_intro1, manipulators_intro2}. Previous work has explored the relationship between expressive robot motion and humans' interpretation of robot movement, resulting in successful nonverbal communication of the robots' intentions \cite{interpret1, interpret2}.\\ While approaches leveraging LMA show substantial results, the prime focus has been on simply expressing affective states (i.e., all kinds of internal states that influence emotion, motivation, and behavior) for the sake of expressivity, rather than expressing affective states to communicate the robot's functional capabilities, which is crucial for human-robot collaboration. Additionally, prior works have prioritized the Laban \textsl{Effort} category while discounting the others (Body, Space, and Shape) due to their impracticality on robots with low degrees of freedom---which greatly curtails the potential expressiveness of the generated behaviors. Finally, the exploration of LMA techniques in expressing emotional and functional trajectories via manipulators is at a nascent stage. We present the following contributions: 
\begin{enumerate}
  \item Enable \textsl{functional expressivity} on robotic manipulators by incorporating a combination of Laban Effort and Laban Shape to generate two variants of a novel expression (Hesitancy), allowing the robot to express the uncertainty of succeeding at its task.
  \item Utilization of Laban Shape alongside Laban Effort to increase the expressivity of emotive manipulator trajectories.
\end{enumerate}
\section{RELATED WORK}
The theory of Laban Movement Analysis (LMA) labels human movement using four categories: Body, Effort, Shape, and Space. While Body relates to the movement of individual body parts, Space indicates where the motion is located within the kinosphere, which refers to the area of potential body movement. The Effort category relates the body's internal intention to its motion attributes, such as strength and timing, and the Shape category focuses on how the body changes its form during movement and its interaction with the surrounding space \cite{laban1975modern, bartenieff2013body}.\\ In the context of expressive robotics, the principal category used is the Laban Effort category, which is further divided into subcategories: Space, Weight, Time, and Flow. Flow defines the body's sense of restriction and freedom, and can be labeled as either Bound or Free Flow. Space defines the body's attitude toward a target, where the motion is either Direct or Indirect. Time classifies the body's movement as either Sudden or Sustained depending on the time taken for traversal. Weight defines how much force is used by the body during movement, which varies from Strong to Light.\\ The next vital category is the Shape category, which is further classified into: Shape Forms, Modes of Shape Change, Shape Qualities, and Shape Flow Support. Shape Forms refers to the static shape that the body takes on. The five primary Shape Forms are Ball-like, Wall-like, Pin-like, Screw-like, and Pyramid-like \cite{shapeforms}. Modes of Shape Change describe the interaction and relationship between the body and the environment and can be further classified into Shape Flow, Directional, and Carving. Shape Qualities describe the way the body is actively changing toward some point in space. A few examples of Shape Quality include Rising, Sinking, Spreading, Enclosing, Advancing, and Retreating. Shape Flow Support describes the way the torso changes in shape to support movements in the rest of the body.\\
To express specific emotions nonverbally, a robot incorporates a fixed combination of these parameters corresponding to each emotion \cite{differential1,humanoid1,manipulator1}. Most of the prior work that leverages LMA for expressive robotics has primarily targeted robots with few degrees of freedom, resulting in the exploitation of only the Laban Effort category. This is because Laban Effort directly influences motion dynamics (e.g., speed, force, and fluidity), which can be effectively represented even with limited joints or actuators. The other Laban categories demand greater physical flexibility and articulation, making them impractical for robots with constrained motion capabilities. Even though the Laban Shape category has been leveraged by humanoids \cite{humanoid1, humanoid2, humanoid3}, it is yet to be applied to other robots.\\ 
Considering the usage of LMA on Unmanned Aerial Vehicles (UAVs), eight unique expressive trajectories authored by a Laban artist were deployed on a quadrotor and the robots' expressions were successfully corroborated via human trials \cite{drone1}. This work was later extended by creating an algorithm to generate task-specific expressive paths for UAVs, which were accurately identified by participants \cite{drone2}. However, these works only employed the Laban Effort category. In \cite{differential1}, the foundation for expressive differential drive robots was set by generating expressive trajectories for six different expressions (Confident, Happy, Sad, Shy, Rushed, and Lackadaisical) via Laban Effort parameters. In further work, LMA was proven to conform with the motion patterns of a differential drive robot without any change in its velocity, showing that the path shape influences peoples' perception of the robots' intention \cite{differential3}. Expressive trajectories led to the differential drive robots being perceived as warm, competent, or uncomfortable depending upon the emotion expressed \cite{differential2}. More recently, a few contributions have extended expressiveness using LMA to humanoids, where a combination of the Space and Shape attributes become relevant. In \cite{humanoid1}, the fusion of both attributes was used on the NAO humanoid robot to represent different ballet characters, which were successfully distinguished by the participants. Additionally, the interpretation of a humanoid’s intent during a handover/handoff task while varying Laban Shape and Effort parameters was analyzed, resulting in the assessment of clarity, friendliness, and predictability for each expressive handoff \cite{humanoid2}.\\
\begin{figure*}[]
 \centering
\includegraphics[width=\linewidth]{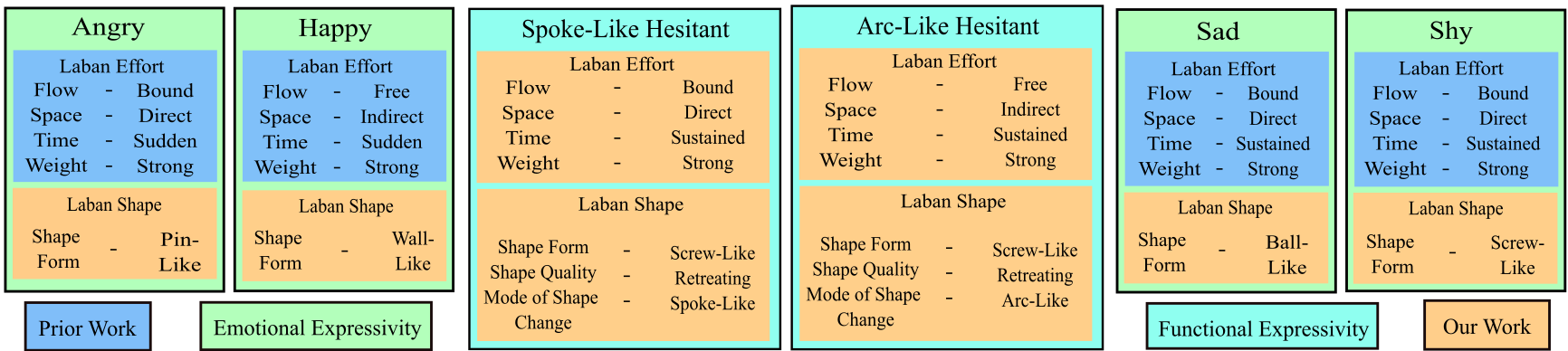}
\caption{For generating expressive motion in robotic manipulators, prior work employs Laban Effort to express five expressions (Angry, Sad, Happy, Shy, and Confident). In our work, we employ Laban Shape alongside Laban Effort to enhance four emotions and facilitate \textsl{functional expressivity} via two variants (Arc-Like and Spoke-Like) of Hesitant trajectories.}
\label{fig:prior_work}
\end{figure*}
Due to regular interactions between humans in everyday life, we tend to relate to a humanoid robots' anthropomorphic form factor and easily interpret its nonverbal cues \cite{anthropomorphic}, which doesn't translate to robotic manipulators, resulting in only two prior works exploring the application of LMA on robotic manipulators. In \cite{manipulator1}, a framework for navigating manipulators through obstacle clutter while being expressive by encoding all of the Laban Effort parameters into a Linear Quadratic (LQ) optimal control problem was developed. While this work enabled manipulators to generate stylized collision-free motions for navigating between configurations, it neither focused on enabling the robot to express a fixed set of predefined emotions nor performed human trials. In \cite{manipulator2}, the Laban Effort category was used to enable a UR5 manipulator to express five expressions (Happy, Sad, Shy, Confident, and Angry). Despite obtaining favorable responses during the human trials, participants were prompted to label the robot with expressions from a given list, thus not encapsulating the true range of human beliefs about expressiveness. Moreover, neither of these works has leveraged the Laban Shape parameters for better expression clarity, which would lead to clearer communication between the robot and the human \cite{new,humanoid1, clarityshape}. Even though all the prior works discussed so far enable robots to express happiness, sadness, anger, confidence, and shyness, none of them have attempted expressing Hesitancy or Uncertainty, which would enable \textsl{functional expressivity} and has high utility and importance in human-robot collaboration \cite{Hesitance, Hesitance1}. Hesitancy in robots is often studied in the context of uncertainty, deliberation, or delayed decision-making, which can make robots appear more human-like and relatable \cite{hesitant1, hesitant2}.\\
\begin{figure*}[]
\centering
\includegraphics[width=\linewidth]{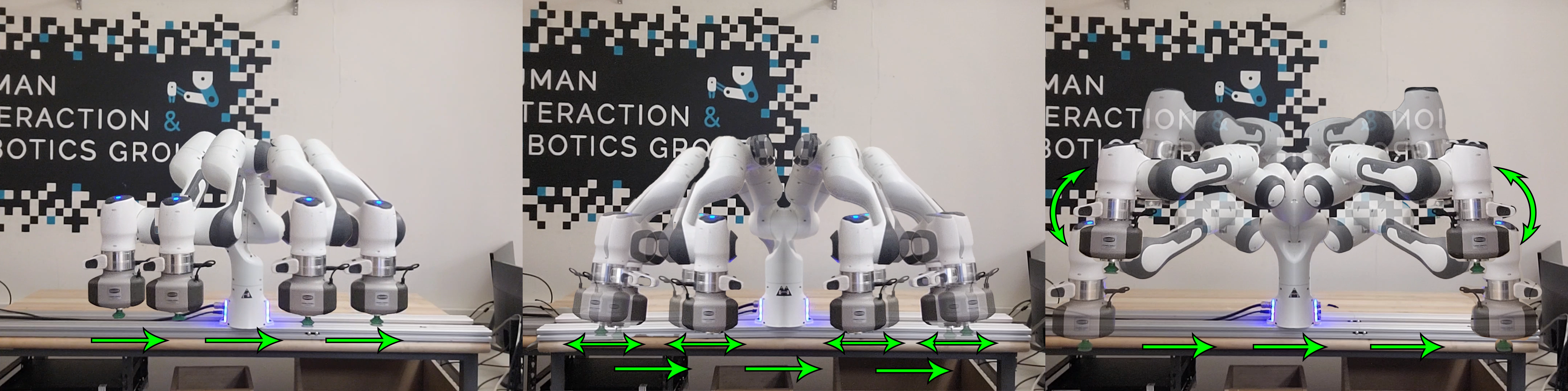}
\caption{A comparison of the Shy trajectory (Left), the Spoke-Like Hesitant trajectory (Middle), and the Arc-Like Hesitant trajectory (Right). The transparent instances of the robot represent the sudden Retreats that encode the uncertain behavior. The Retreats are characterized by the Laban Shape Quality, which constitutes how the robot is actively changing toward its desired waypoint.}
\label{fig:all_three}
\end{figure*}
As seen in Fig. \ref{fig:prior_work}, our work leverages the combination of Laban Effort and Laban Shape while applying it to robotic manipulators, resulting in trajectories that express specified expressions.
Additionally, we show how to utilize the LMA parameters to generate trajectories that express Hesitancy, paving the way for \textsl{functional expressivity}, where we enable the robot to communicate its abilities and the likelihood of its success while performing a task. Finally, our human trials allow the participants to freely label the trajectories without providing pre-selected options, resulting in an unbiased evaluation of the expressivity of the robot with a more robust understanding of how humans describe expressive robot motion.

\section{METHODS}
Prior works have only utilized the Laban Effort category for generating expressive emotive trajectories on robotic manipulators.
We use the Laban Effort parameter definitions given by \cite{manipulator1}, which is the state-of-the-art for generating LMA-based expressions on manipulators. The Laban Effort parameter definitions that are specifically defined for expressive manipulator trajectories are as follows:

\begin{enumerate}
  \item Time: Refers to the time the robot will take to complete the trajectory. Sustained Time is when the robot takes a long time to finish the traversal whereas Sudden is when the robot quickly executes the motion.
  \item Space: Focuses on the global trajectory of the robot. If the robot pursues a straight line, it is referred to as Direct whereas if it executes a curve, the result is Indirect.
  \item  Flow: Determines the flexibility of the trajectory using a fixed number of waypoints the robot must pass through. If the number of waypoints equals 2, it would lead to Bound Flow. If the number of waypoints exceeds 2, it results in Free Flow.
  \item  Weight: Signifies the stiffness of the joints; if the Weight is Light, the joints on the end effector will be actuating freely while in the case of Strong Weight, the end effector joints will not undergo any motion.
\end{enumerate} 
To ensure the generalizability of our approach across manipulators, we use Strong Laban Weights for all expressions. Every other Effort parameter value corresponding to each of the four emotions (Shy, Angry, Sad, and Happy) was the same as in \cite{manipulator1}:
\begin{enumerate}
  \item  Shy: Sustained Time, Bound Flow, and Direct Space.
  \item  Angry: Sudden Time, Bound Flow, and Direct Space.
  \item  Sad: Sustained Time, Bound Flow, and Direct Space.
  \item  Happy: Sudden Time, Free Flow, and Indirect Space.
\end{enumerate} 
We add the usage of Laban Shape alongside the Laban Effort parameters to increase the clarity of expressions for four emotions (Happy, Sad, Angry, and Shy). Specifically, we leverage Laban Shape Forms, which refers to the static shapes that the robot body takes throughout its motion from one location to another. Since a 7-DOF robotic manipulator is holonomic (i.e., its controllable degrees of freedom are equal to its total degrees of freedom), it can traverse between two different end-effector poses while maintaining a static shape. The proposed framework is meant to be portable to any robotic manipulator with similar characteristics. The various Shape Forms corresponding to the expressions are shown in Fig \ref{fig:all_shapes_together}. The description of the Laban Shape Form for each expression is as follows:

\begin{enumerate}
  \item  Happy: The manipulator stretches itself upwards while the end-effector maintains a 45$\degree$ angle between the horizontal and vertical. This originates from the Wall-like Laban Shape Form, which signifies stability and strength.
  \item  Sad: Starting from the robots' base, every consecutive joint curves inwards to form an approximate semicircle. This resembles the Ball-like Laban Shape Form, which indicates a sense of vulnerability and withdrawal.
  \item  Shy: The robot leans forward without significant curvature between the joints. However, the end-effector faces downwards. This emanates from the Screw-like Laban Shape Form, which conveys anxiety and uneasiness.
  \item  Angry: The robot stretches forward towards the end of the kinosphere while the end-effector maintains a $45\degree$ angle between the horizontal and vertical. This is derived from the Pin-like Laban Shape Form, which provides a sense of tension and aggression.
\end{enumerate}
Instead of relying on the frequently utilized confident expression, we enable \textsl{functional expressivity} on the manipulator by introducing two variants of Hesitancy; Spoke-like and Arc-like. In LMA theory, Spoke-like and Arc-like are the two types of Directional Modes of Shape Change,  which describe the way the body interacts with the environment and the relationship between them. In Spoke-Like movement, the body follows a linear, straight path. In Arc-like motion, the body follows a curved trajectory, like an arc. In addition to Modes of Shape Change, we also leverage Shape Quality, which refers to the way the body is actively changing
toward some point in space. Specifically, we leverage the Retreating Shape Quality, which indicates the intent of the robot to backtrack regularly instead of confidently committing to its initial direction, as seen in Fig. \ref{fig:all_three}. In Spoke-Like Hesitancy, the movement follows a direct, linear path between the waypoints, with intermittent linear Retreats in the opposite direction. In Arc-like Hesitancy, the movement follows a curved path, with intermittent arc-like Retreats in the opposite direction. The formulated Laban Effort Parameters for both variants are as follows:
\begin{enumerate}
  \item Time: Sustained (Both)
  \item Flow: Bound (Spoke-Like) and Free (Arc-Like)
  \item Space: Direct (Spoke-Like) and Indirect (Arc-Like)
  \item Weight: Strong (Both)
\end{enumerate}
The Laban Shape Form used to describe Hesitancy and Shyness is the same since both expressions have similar characteristics (i.e., high level of caution and reluctance to engage). The difference between the Spoke-Like Hesitancy and Shyness is the Shape Quality, while the Shape Forms and the Laban Effort parameters are the same. In Arc-Like Hesitancy, the Flow and Space Effort parameters change to account for the curvy path.
\begin{figure}
\centering
\includegraphics[width=\linewidth]{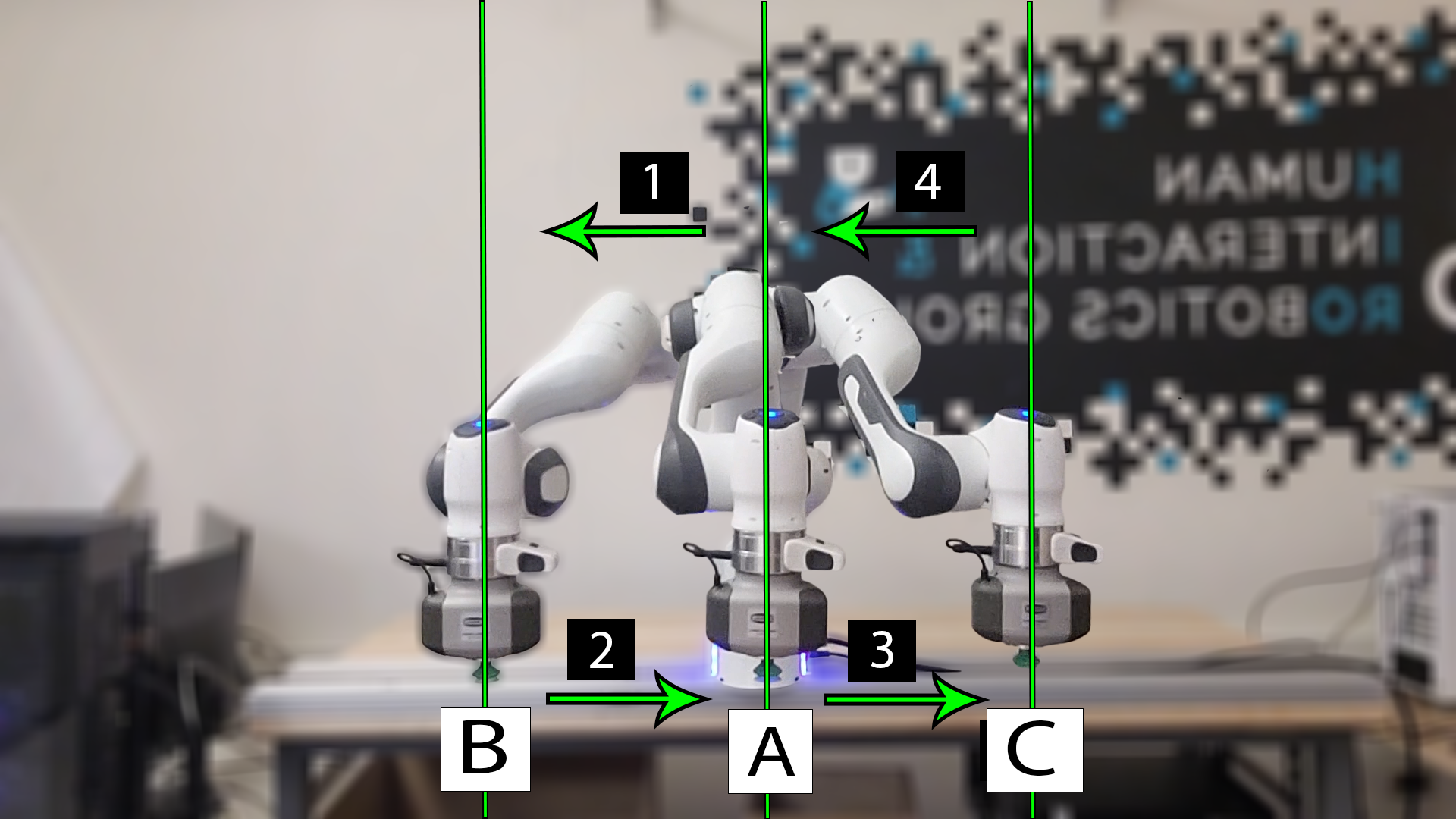}
\caption{For all the expressive trajectories, the end effector traverses through a point within the lines A, B, and C in a fixed order. While recording each trajectory, the robot was actuated through these waypoints while satisfying the Laban constraints. Afterwards, each path corresponding to the expression was replayed from its initial configuration.}
\label{fig:waypoints}
\end{figure}
\begin{figure*}[hbt!]
\begin{subfigure}{.32\linewidth}
  \includegraphics[width=\linewidth]{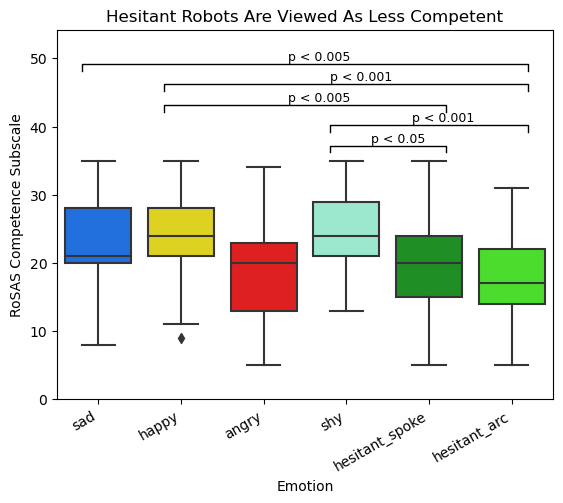}
  \caption{Hesitant trajectories significantly reduce the competence of the robot while happiness and Shyness fare well.}
\end{subfigure}\hfill 
\begin{subfigure}{.32\linewidth}
  \includegraphics[width=\linewidth]{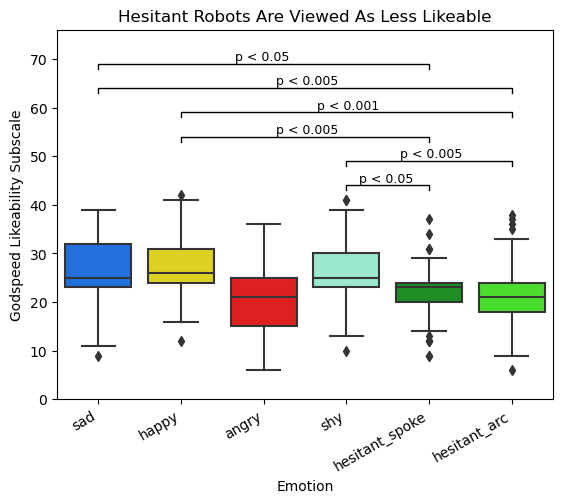}
  \caption{Hesitant and Angry expressions result in low likeability while happiness and Shyness flourish.}
  \label{energydetPSK}
\end{subfigure}\hfill
\begin{subfigure}{.32\linewidth}
  \includegraphics[width=\linewidth]{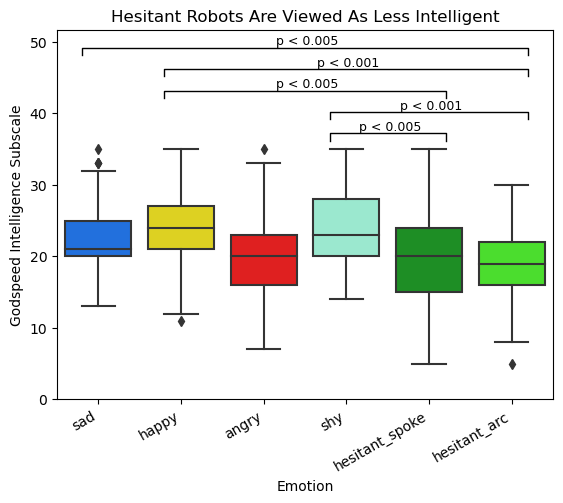}
  \caption{Hesitant and Angry robots are not considered intelligent while Happy and Shy robots are.}
  \label{velcomp}
\end{subfigure}
\caption{A comparison of the competence, likeability and intelligence RoSAS subscale values with all six trajectories. The figures show that Arc-Like Hesitant and Spoke-Like Hesitant are perceived to be less competent than every emotion except Angry, which fares similarly.}
\label{fig:RoSAS}
\end{figure*}
\section{EXPERIMENTAL DESIGN}
An IRB-approved human subjects study was conducted to validate the use of Laban Shape in enhancing the expressivity of the Happy, Sad, Angry, and Shy expressions, as well as the interpretation of the two variants of Hesitancy. We generate an expressive trajectory on the Franka Emika Panda for each of the six expressions: Sad, Happy, Angry, Shy, Arc-Like Hesitant, and Spoke-Like Hesitant. In each of these trajectories, the robot's end effector traverses through waypoints that lie on 3 fixed lines -- A, B, and C, as seen in Fig. \ref{fig:waypoints}. The trajectory begins at the initial waypoint that lies on line A. Then, the robot travels to reach the second waypoint, which lies on line B. Eventually, it travels to the third waypoint, which lies on line C before returning to its initial position. All the trajectories were recorded by actuating the robot sequentially from lines A to B, from B to C and finally, back to A. This was performed while satisfying all the Laban Effort and Shape constraints. Once recorded, the MoveIt motion planning framework was used to navigate the robot to the initial configuration of the expressive trajectory that lies on the line A, and the series of recorded joint angles were replayed in order until the entire trajectory was executed. \\
\subsection{Protocol}
After completing consent forms, participants were required to read instructions about the survey before attempting it. Then, each participant was asked questions regarding their attitude towards robots, which were taken from the Negative Attitudes Towards Robots Scale (NARS) \cite{NARS}. Afterward, a video of the robot executing a randomly selected expressive trajectory was played to the participants, following which they were asked to name the top three expressions they perceived the robot to be conveying. They were then asked questions from the Competence Subscale of the Robotic Social Attributes Scale (RoSAS) and the Intelligence and Likeability Subscales of the Godspeed Questionnaire \cite{ROSAS, godspeed}. This was repeated six times until participants had viewed and assessed all emotive trajectories. 
\subsection{Hypotheses}
Through this study, we investigated the following hypotheses:
\begin{itemize}
  \item $H_1$: Participants will find the Hesitant variants to be less competent than all the emotions.
  \item $H_2$: Participants will find no difference in terms of competence, likeability and intelligence between the two variants of Hesitant trajectories whilst clearly distinguishing them from the Shy Trajectory.
  \item $H_3$: Participants will be able to distinguish the enhanced emotive expressions across at least one axis of the VAD scale despite allowing them to freely describe the expression.
\end{itemize}
The three hypothesis were established before conducting the experiment. We established H1 based on the assumption that a hesitant entity inherently exhibits a lack of competitiveness, stemming from reduced confidence, certainty, and prowess. We developed H2 with two logical premises: (1) that both variants of hesitancy would be perceived similarly by participants, despite differing in Laban Shape qualities; and (2) that the Hesitant trajectory would be perceptually distinguishable from the Shy trajectory, even though both share the same Laban Effort values. Lastly, we designed H3 to highlight a clear distinction in participants' perception of each of the robot’s trajectories.
\section{RESULTS}
\begin{figure}[hbt!]
\centering
\includegraphics[width=\linewidth]{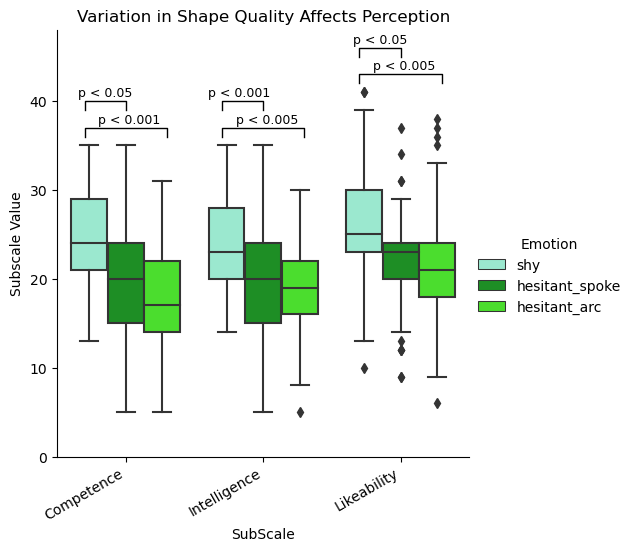}
\caption{The Shy expression was perceived to be more competent (p=0.0077), intelligent (p=0.0044) and likable (p=0.0220) than the Spoke-Like Hesitant expression despite the Laban Shape Quality being the only difference. Despite the change in Laban Effort parameters, both variants fared similarly in all three metrics with Arc-Like motion being less competent (p=0.004), likable (p=0.0031) and intelligent (p=1.72e-05) than the Shy motion.}
\label{fig:shape_variance}
\end{figure}
\begin{figure*}[hbt!]
\begin{subfigure}{.32\linewidth}
  \includegraphics[width=\linewidth]{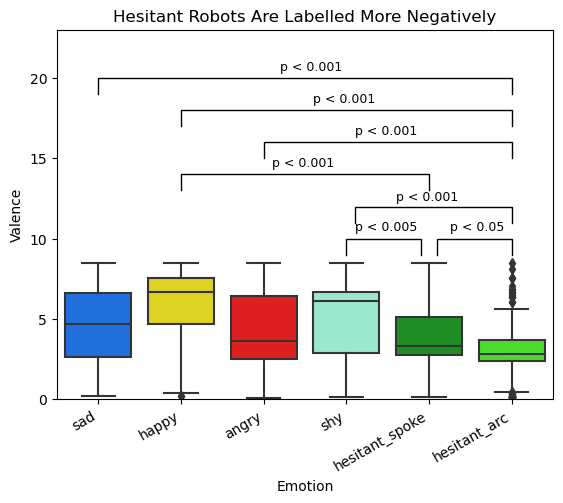}
  \caption{The Happy trajectory was found to be very positive with high valence while the Hesitant variants performed poorly.}
\end{subfigure}\hfill 
\begin{subfigure}{.32\linewidth}
  \includegraphics[width=\linewidth]{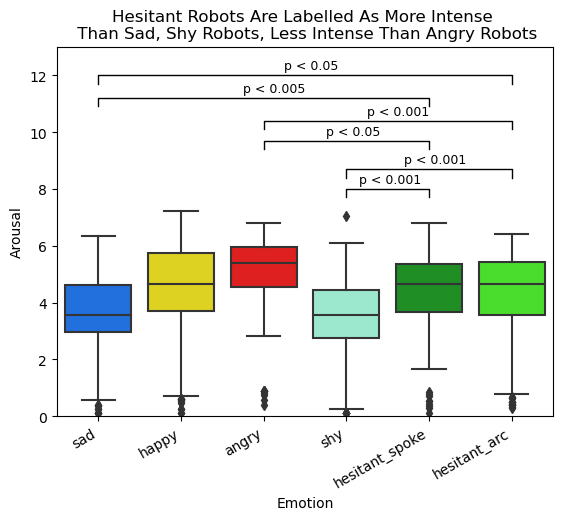}
  \caption{The Angry path was perceived to present the highest arousal value while the Shy path was perceived to be the least arousing.}
  \label{energydetPSK}
\end{subfigure}\hfill
\begin{subfigure}{.32\linewidth}
  \includegraphics[width=\linewidth]{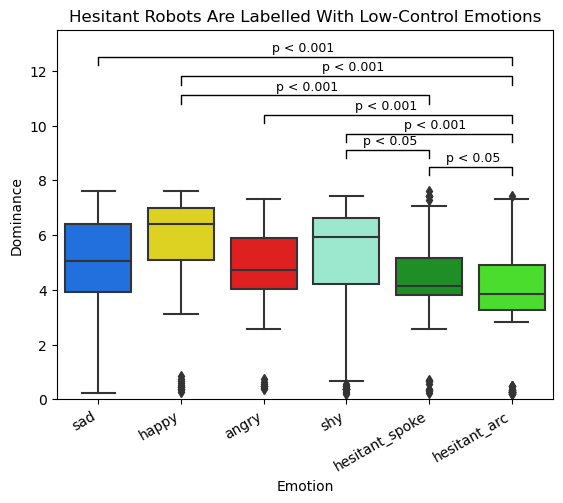}
  \caption{The Happy trajectory resulted in the highest dominance while the Hesitant variants were found to be the least dominant.}
  \label{velcomp}
\end{subfigure}
\caption{A comparison of the Valence, Arousal, and Dominance values for the user input with all six trajectories shows that every expression has been distinguished by the participants across at least one of the VAD axes when they freely labeled the expression.}
\label{fig:vad}
\end{figure*}
In total, 50 participants (Ages 19-76, M=23, F=26, and U=1), were recruited via Prolific, an online research platform that enables data collection by connecting diverse participants with researchers. One participant was removed for failure to follow directions. All data was analyzed using Tukey's Honestly Significant Difference (HSD) test after we performed an ANOVA (Analysis of Variance test), which resulted in the differences in means being statistically significant. No multimodalities were observed within the data. As seen in Fig. \ref{fig:RoSAS}, the comparison of the RoSAS metrics with each expressive motion provided the following results:
\begin{enumerate}
    \item Competence: The participants found the Hesitant expressions to be less competent than Sad (p=0.0009,3.59e-07), Happy (p=3.5e-07, 0.0008), and Shy (p=5.2e-05,1.8e-07) expressions. There is minimal difference in the competence between the Hesitant and Angry trajectories. This is in line with the notion of incapability affiliated with Uncertainty.
  \item Likeability: The Arc-Like and Spoke-like variants were less likable than the Happy (p=1.84e-05,1.9e-05), Sad (p=9.8e-05, 0.0009), and Shy (p=0.0004, 0.00095) expressions and they fared similarly to the Angry expression. This validates the negative connotation associated with Uncertainty.
  \item Intelligence: The Hesitant trajectory was perceived to be less intelligent compared to the Happy (p=2.85e-07,0.00037), Sad (p=0.00074), and Shy ones (p=5.79e-07,0.0007). The perception was very similar between the Hesitant expression and the Angry expression. This corroborates the lack of intelligence affiliated with Uncertainty.
\end{enumerate}
This finding \textbf{partially validates $H_1$}, confirming that both variants of Hesitant motion result in the robot being perceived as less competent compared to the Happy, Sad, and Shy expressions. The Angry trajectory was perceived with similar negativity, which is acceptable due to the lack of applicability of this emotion in human-robot collaboration.\\ As seen in Fig. \ref{fig:shape_variance}, another major finding is the high similarity of all the RoSAS subscale metrics between the Arc-Like and Spoke-Like Hesitancy. Despite the change in Laban Effort parameters (Space and Flow), we believe that the consistency in the Laban Shape Forms and the Laban Shape Quality resulted in participants perceiving both variants as equivalent expressions with similar competence, intelligence, and likeability with a lack of significance between them. Despite the Shy trajectory and Spoke-Like Hesitant trajectory having the same Laban Effort parameter values and Laban Shape Forms, we believe that the addition of the Retreating Laban Shape Quality resulted in a significant difference between the expressions with the Shy motion being perceived as more competent (p=0.0077), more intelligent (p=0.0044), and more likable (p=0.022). This finding \textbf{validates $H_2$}, which indicates that changes in Laban Shape values have a higher impact on the robot's likability, intelligence, and competence compared to changes in the Laban Effort values.  

We use Valence-Arousal-Dominance (VAD) values to quantify and analyze the words that the participants used to label the robot's expressions, a novel approach for examining human impressions of expressive robots. The VAD Graph is a three-dimensional model used to represent and quantify expressions based on three key factors: Valence, Arousal, and Dominance. Valence represents the positivity or negativity of an expression. Arousal quantifies the intensity or energy level of an expression. Finally, Dominance indicates the level of control or power associated with an expression. The VAD model is particularly useful since expressions are not binary, and this approach allows for a continuous, multidimensional representation of them rather than fixed categories. After analyzing the data by referring to a pre-existing mapping of words to their corresponding VAD values \cite{vad_dataset}, the following finding is observed:
\begin{enumerate}
  \item Valence: The Happy trajectory was found to be perceived more positively than all the other expressions. On the other hand, Arc-Like Hesitant was perceived more negatively than all the others. Interestingly, the Shy trajectory was perceived very positively compared to the Spoke-Like (p=0.0032) and Arc-Like (p=1.04e-11) Hesitant one. The results show that the positive expressions (i.e., Happy and Shy) obtained higher valence values, corroborating the utilization of Laban Shape Forms.
  \item Arousal: The Angry trajectory proved to be more intense than all the other expressions. The Sad expression was milder than Happy(p=0.00029), Spoke-Like Hesitant (p=0.0007), and Arc-Like Hesitant (p=0.032). The Shy movement was found to be the least intense. Notably, the Hesitant variants were found to be significantly more arousing than the Shy motion (p=0.0013, 9.53e-07). Thus, the participants were able to clearly distinguish the arousal aspect of the trajectories while finding both variants of Hesitant to be on par with each other.
  \item Dominance: The Happy trajectory was found to be more dominant than all the other expressions. Arc-Like Hesitant was less dominant than all the others. Interestingly, Shy was found to be significantly more dominant than Spoke-Like Hesitant (p=0.0067) and Arc-Like Hesitant (p=1.5e-09). There is a clear distinction between every expression's dominant nature. We believe this validates the uncertain nature of the Retreating Shape Quality and the Screw-Like Laban Shape Form. 
  
\end{enumerate}
As seen in Fig. \ref{fig:vad}, this finding \textbf{validates $H_3$}, where all of the labeled expressions were visibly different along at least one axis of the VAD scale due to the enhancement of the expressivity.
Participants found both variants of the Hesitant trajectory to be less competent than every other expression except the Angry one, which fared similarly. This is aligned with the notion of incapability associated with the term "Uncertainty". Additionally, participants found minimal difference between the two variants of Hesitant motion in terms of competence, likeability, and intelligence, while distinguishing them from the Shy expression. This indicates that variation in Laban Shape parameters affects greater change in participant perceptions than variations in Laban Effort, validating the usage of Laban Shape as a more effective driver of robot expressivity, thus boosting human-robot collaboration via \textsl{functional expressivity}. Finally, the participants were able to distinguish the enhanced emotive expressions across at least one axis of the VAD scale, validating the efficacy of adding Laban Shape Forms to the base expressions.

\section{CONCLUSION AND FUTURE WORK}
This paper aims to improve transparency and alignment in human-robot collaboration by enabling \textsl{functional expressivity} on robotic manipulators, which entails expressing the robot's abilities and the likelihood of its success while performing a task. This is achieved by employing Laban Shape alongside Laban Effort to express "Uncertainty" via two variants of Hesitant trajectories (Arc-Like and Spoke-Like). Through a human subjects study, we show that both variants of the Hesitant trajectory are perceived to be less competent than other expressions, which validates the functional incapability of uncertain behaviors. The two variants of Hesitant motion were also found to be perceived similarly in terms of competence, likeability, and intelligence. We use the same set of metrics to show that Spoke-Like Hesitance is perceived differently from Shy despite the only difference being the Retreating Laban Shape Quality. This highlights the utility of Laban Shape as a strong driver of emotive expression in manipulators. Participants were also able to distinguish the enhanced emotive expressions across at least one axis of the VAD scale, substantiating the usage of Laban Shape Forms. A limitation of this work is the usage of VAD values for assessing \textsl{functional expressivity} even though it represents affective states instead of just basic emotions. Future work can explore this avenue and also focus on the applications of this research in a collaborative setting, where a human and a robot can attempt to complete a task together while the effects of \textsl{functional expressivity} are analyzed within the context of a human-robot dyad. Specifically, we can explore leveraging hesitation to balance the human's expectation on the robot in high-risk scenarios like transferring sharp or fragile objects and performing tasks that are prone to failure like navigating through immense clutter of obstacles without colliding with them.
\section{ACKNOWLEDGMENTS}
This work was supported by the Army Research Laboratory under the Grant \# W911NF-21-2-02905.
\bibliographystyle{plain}
\bibliography{references}

\begin{thebibliography}{10}

\bibitem{Hesitance1}
Henny Admoni, Anca Dragan, Siddhartha~S Srinivasa, and Brian Scassellati.
\newblock Deliberate delays during robot-to-human handovers improve compliance with gaze communication.
\newblock In {\em Proceedings of the 2014 ACM/IEEE international conference on Human-robot interaction}, pages 49--56, 2014.

\bibitem{vad_dataset}
Bagus~Tris Atmaja and Masato Akagi.
\newblock Deep learning-based categorical and dimensional emotion recognition for written and spoken text.
\newblock 2019.

\bibitem{interpret2}
Alexandra Bacula.
\newblock Socially communicative multi-robot motion and formation: Algorithm development and validation.
\newblock 2023.

\bibitem{differential_drive_intro}
Alexandra Bacula and Amy LaViers.
\newblock Character design and validation on aerial robotic platforms using laban movement analysis.
\newblock In {\em Social Robotics: 10th International Conference, ICSR 2018, Qingdao, China, November 28-30, 2018, Proceedings 10}, pages 202--212. Springer, 2018.

\bibitem{humanoid1}
Alexandra Bacula and Amy LaViers.
\newblock Character synthesis of ballet archetypes on robots using laban movement analysis: comparison between a humanoid and an aerial robot platform with lay and expert observation.
\newblock {\em International Journal of Social Robotics}, 13(5):1047--1062, 2021.

\bibitem{bartenieff2013body}
Irmgard Bartenieff and Dori Lewis.
\newblock {\em Body movement: Coping with the environment}.
\newblock Routledge, 2013.

\bibitem{bernardet2019assessing}
Ulysses Bernardet, Sarah Fdili~Alaoui, Karen Studd, Karen Bradley, Philippe Pasquier, and Thecla Schiphorst.
\newblock Assessing the reliability of the laban movement analysis system.
\newblock {\em PloS one}, 14(6):e0218179, 2019.

\bibitem{bernstein2015laban}
Ran Bernstein, Tal Shafir, Rachelle Tsachor, Karen Studd, and Assaf Schuster.
\newblock Laban movement analysis using kinect.
\newblock {\em International Journal of Computer and Information Engineering}, 9(6):1567--1571, 2015.

\bibitem{nonverbal}
Cynthia Breazeal, Cory~D Kidd, Andrea~Lockerd Thomaz, Guy Hoffman, and Matt Berlin.
\newblock Effects of nonverbal communication on efficiency and robustness in human-robot teamwork.
\newblock In {\em 2005 IEEE/RSJ international conference on intelligent robots and systems}, pages 708--713. IEEE, 2005.

\bibitem{burgoon2011nonverbal}
Judee~K Burgoon, Laura~K Guerrero, and Valerie Manusov.
\newblock Nonverbal signals.
\newblock {\em Handbook of interpersonal communication}, pages 239--280, 2011.

\bibitem{ROSAS}
Colleen~M Carpinella, Alisa~B Wyman, Michael~A Perez, and Steven~J Stroessner.
\newblock The robotic social attributes scale (rosas) development and validation.
\newblock In {\em Proceedings of the 2017 ACM/IEEE International Conference on human-robot interaction}, pages 254--262, 2017.

\bibitem{humanoid3}
Ramya Challa, Luke Sanchez, Cristina~G Wilson, and Heather Knight.
\newblock Take it! exploring cartesian features for expressive arm motion.
\newblock In {\em 2024 33rd IEEE International Conference on Robot and Human Interactive Communication (ROMAN)}, pages 703--710. IEEE, 2024.

\bibitem{interpret1}
Victoria Chen, Yao-Lin Tsai, and Heather Knight.
\newblock Determining success and attributes of various feeding approaches with a mobile robot.
\newblock In {\em 2022 17th ACM/IEEE International Conference on Human-Robot Interaction (HRI)}, pages 713--717. IEEE, 2022.

\bibitem{humanoid2}
Stone Cheng.
\newblock Study of emotion rendering design for humanoid robots compiled with real-time music mood perception.
\newblock In {\em 2017 26th IEEE International Symposium on Robot and Human Interactive Communication (RO-MAN)}, pages 647--652. IEEE, 2017.

\bibitem{drone1}
Hang Cui, Catherine Maguire, and Amy LaViers.
\newblock Laban-inspired task-constrained variable motion generation on expressive aerial robots.
\newblock {\em Robotics}, 8(2):24, 2019.

\bibitem{differential2}
Ebru Emir and Catherine~M Burns.
\newblock Evaluation of expressive motions based on the framework of laban effort features for social attributes of robots.
\newblock In {\em 2022 31st IEEE International Conference on Robot and Human Interactive Communication (RO-MAN)}, pages 1548--1553. IEEE, 2022.

\bibitem{mataric}
Thomas Groechel, Zhonghao Shi, Roxanna Pakkar, and Maja~J Matari{\'c}.
\newblock Using socially expressive mixed reality arms for enhancing low-expressivity robots.
\newblock In {\em 2019 28th IEEE International Conference on Robot and Human Interactive Communication (RO-MAN)}, pages 1--8. IEEE, 2019.

\bibitem{manipulator1}
Roshni Kaushik, Anant~Kumar Mishra, and Amy LaViers.
\newblock Feasible stylized motion: Robotic manipulator imitation of a human demonstration with collision avoidance and style parameters in increasingly cluttered environments.
\newblock In {\em Proceedings of the 7th International Conference on Movement and Computing}, pages 1--8, 2020.

\bibitem{functional_new}
Dacher Keltner and Jonathan Haidt.
\newblock Social functions of emotions at four levels of analysis.
\newblock {\em Cognition \& Emotion}, 13(5):505--521, 1999.

\bibitem{differential1}
Heather Knight and Reid Simmons.
\newblock Expressive motion with x, y and theta: Laban effort features for mobile robots.
\newblock In {\em The 23rd IEEE international symposium on robot and human interactive communication}, pages 267--273. IEEE, 2014.

\bibitem{clarityshape}
Heather Knight and Reid Simmons.
\newblock Laban head-motions convey robot state: A call for robot body language.
\newblock In {\em 2016 IEEE international conference on robotics and automation (ICRA)}, pages 2881--2888. IEEE, 2016.

\bibitem{differential3}
Heather Knight, Ravenna Thielstrom, and Reid Simmons.
\newblock Expressive path shape (swagger): Simple features that illustrate a robot's attitude toward its goal in real time.
\newblock In {\em 2016 IEEE/RSJ International Conference on Intelligent Robots and Systems (IROS)}, pages 1475--1482. IEEE, 2016.

\bibitem{manipulators_intro1}
Minae Kwon, Sandy~H Huang, and Anca~D Dragan.
\newblock Expressing robot incapability.
\newblock In {\em Proceedings of the 2018 ACM/IEEE International Conference on Human-Robot Interaction}, pages 87--95, 2018.

\bibitem{manipulator2}
Carlo La~Viola, Laura Fiorini, Gianmaria Mancioppi, Jaeseok Kim, and Filippo Cavallo.
\newblock Humans and robotic arm: Laban movement theory to create emotional connection.
\newblock In {\em 2022 31st IEEE International Conference on Robot and Human Interactive Communication (RO-MAN)}, pages 566--571. IEEE, 2022.

\bibitem{laban1975modern}
Raban Laban.
\newblock Modern educational dance.
\newblock {\em MacDonald \& Evans}, 1975.

\bibitem{manipulators_intro2}
Gregory Lemasurier, Gal Bejerano, Victoria Albanese, Jenna Parrillo, Holly~A Yanco, Nicholas Amerson, Rebecca Hetrick, and Elizabeth Phillips.
\newblock Methods for expressing robot intent for human--robot collaboration in shared workspaces.
\newblock {\em ACM Transactions on Human-Robot Interaction (THRI)}, 10(4):1--27, 2021.

\bibitem{shapeforms}
Vera Maletic.
\newblock {\em Body, space, expression: The development of Rudolf Laban's movement and dance concepts}, volume~75.
\newblock Walter de Gruyter, 1987.

\bibitem{hesitant1}
AJung Moon, Chris~AC Parker, Elizabeth~A Croft, and HF~Machiel Van~der Loos.
\newblock Did you see it hesitate?-empirically grounded design of hesitation trajectories for collaborative robots.
\newblock In {\em 2011 IEEE/RSJ International Conference on Intelligent Robots and Systems}, pages 1994--1999. IEEE, 2011.

\bibitem{hesitant2}
AJung Moon, Chris~AC Parker, Elizabeth~A Croft, and HF~Machiel Van~der Loos.
\newblock Design and impact of hesitation gestures during human-robot resource conflicts.
\newblock {\em Journal of Human-Robot Interaction}, 2(3):18--40, 2013.

\bibitem{psych}
Helen Payne.
\newblock The psycho-neurology of embodiment with examples from authentic movement and laban movement analysis.
\newblock {\em American Journal of Dance Therapy}, 39:163--178, 2017.

\bibitem{new}
Ali-Akbar Samadani, Sarahjane Burton, Rob Gorbet, and Dana Kulic.
\newblock Laban effort and shape analysis of affective hand and arm movements.
\newblock In {\em 2013 Humaine Association conference on affective computing and intelligent interaction}, pages 343--348. IEEE, 2013.

\bibitem{anthropomorphic}
Shane Saunderson and Goldie Nejat.
\newblock How robots influence humans: A survey of nonverbal communication in social human--robot interaction.
\newblock {\em International Journal of Social Robotics}, 11(4):575--608, 2019.

\bibitem{drone2}
Megha Sharma, Dale Hildebrandt, Gem Newman, James~E Young, and Rasit Eskicioglu.
\newblock Communicating affect via flight path exploring use of the laban effort system for designing affective locomotion paths.
\newblock In {\em 2013 8th ACM/IEEE International Conference on Human-Robot Interaction (HRI)}, pages 293--300. IEEE, 2013.

\bibitem{uav_intro}
David St-Onge, Ulysse C{\^o}t{\'e}-Allard, Kyrre Glette, Benoit Gosselin, and Giovanni Beltrame.
\newblock Engaging with robotic swarms: Commands from expressive motion.
\newblock {\em ACM Transactions on Human-Robot Interaction (THRI)}, 8(2):1--26, 2019.

\bibitem{NARS}
Dag~Sverre Syrdal, Kerstin Dautenhahn, Kheng~Lee Koay, and Michael~L Walters.
\newblock The negative attitudes towards robots scale and reactions to robot behaviour in a live human-robot interaction study.
\newblock {\em Adaptive and emergent behaviour and complex systems}, 2009.

\bibitem{functionalexpressivity1}
Daniel Szafir, Bilge Mutlu, and Terry Fong.
\newblock Communicating directionality in flying robots.
\newblock In {\em Proceedings of the Tenth Annual ACM/IEEE International Conference on Human-Robot Interaction}, pages 19--26, 2015.

\bibitem{tabrez2020survey}
Aaquib Tabrez, Matthew~B Luebbers, and Bradley Hayes.
\newblock A survey of mental modeling techniques in human--robot teaming.
\newblock {\em Current Robotics Reports}, 1:259--267, 2020.

\bibitem{tsachor2017somatic}
Rachelle~P Tsachor and Tal Shafir.
\newblock A somatic movement approach to fostering emotional resiliency through laban movement analysis.
\newblock {\em Frontiers in human neuroscience}, 11:410, 2017.

\bibitem{Hesitance}
Rik Van~den Brule, Gijsbert Bijlstra, Ron Dotsch, Pim Haselager, and Dani{\"e}l~HJ Wigboldus.
\newblock Warning signals for poor performance improve human-robot interaction.
\newblock {\em Journal of Human-Robot Interaction}, 5(2):69--89, 2016.

\bibitem{functionalexpressivity2}
Michael Walker, Hooman Hedayati, Jennifer Lee, and Daniel Szafir.
\newblock Communicating robot motion intent with augmented reality.
\newblock In {\em Proceedings of the 2018 ACM/IEEE International Conference on Human-Robot Interaction}, pages 316--324, 2018.

\bibitem{godspeed}
Astrid Weiss and Christoph Bartneck.
\newblock Meta analysis of the usage of the godspeed questionnaire series.
\newblock In {\em 2015 24th IEEE international symposium on robot and human interactive communication (RO-MAN)}, pages 381--388. IEEE, 2015.

\end{thebibliography}

\end{document}